%% file: main.tex
\documentclass[conf]{new-aiaa} %for conference papers

\usepackage[utf8]{inputenc}
\usepackage{textcomp}
\def\ACAS{$\text{ACAS X}$~}
\def\ACASA{$\text{ACAS Xa}$~}
\def\ACASU{$\text{ACAS Xu}$~}
\def\ACASSU{$\text{ACAS sXu}$~}
\def\ACASR{$\text{ACAS Xr}$~}
\def\TCAS{$\text{TCAS II}$~}

\usepackage{graphicx}
\usepackage{amsmath}
\usepackage[version=4]{mhchem}
\usepackage{siunitx}
\usepackage{longtable,tabularx}
\usepackage{cleveref}

\usepackage{booktabs}
\usepackage{aircraftshapes}

\usepackage{pgfplots}
\usepgfplotslibrary{groupplots}

\pgfplotsset{compat=newest}
\pgfplotsset{every axis legend/.append style={legend cell align=left}}

\pgfplotsset{every axis/.append style={
                    title style={font=\small},
                    tick label style={font=\footnotesize}  
                    }}
\pgfplotsset{every axis label/.style={font=\small}}                    
\pgfplotsset{
legend image code/.code={
\draw[mark repeat=2,mark phase=2]
plot coordinates {
(0cm,0cm)
(0.15cm,0cm)        %% default is (0.3cm,0cm)
(0.3cm,0cm)         %% default is (0.6cm,0cm)
};%
}
}

\definecolor{pastelMagenta}{HTML}{FF48CF}
\definecolor{pastelPurple}{HTML}{8770FE}
\definecolor{pastelBlue}{HTML}{1BA1EA}
\definecolor{pastelSeaGreen}{HTML}{14B57F}
\definecolor{pastelGreen}{HTML}{3EAA0D}
\definecolor{pastelOrange}{HTML}{C38D09}
\definecolor{pastelRed}{HTML}{F5615C}
\definecolor{lightGray}{RGB}{169, 169, 169}
\definecolor{darkRed}{RGB}{139, 0, 0}

\usetikzlibrary{plotmarks}

\newcommand{\argmax}{\operatornamewithlimits{arg\,max}}

\title{Collision Risk and Operational Impact of Speed Change Advisories as Aircraft Collision Avoidance Maneuvers} % Change this :) % Gave this a try....lets see if people like it (LEA)

\author{Sydney M. Katz\footnote{Graduate Student, Department of Aeronautics and Astronautics, Stanford CA} and Mykel J. Kochenderfer\footnote{Associate Professor,  Department of Aeronautics and Astronautics, Stanford CA, AIAA Associate Fellow}}
\affil{Stanford University, Stanford, CA, 94305, USA}
\author{Luis E. Alvarez\footnote{Technical Staff, Surveillance Systems, Lexington MA, AIAA Senior Member}, Michael Owen\footnote{Technical Staff, Surveillance Systems, Lexington MA, AIAA Senior Member}, Samuel Wu\footnote{Associate Staff, Surveillance Systems, Lexington MA}, Marc Brittain\footnote{Technical Staff, Surveillance Systems, Lexington MA, AIAA Senior Member}, and Anshuman Das\footnote{Associate Staff, Surveillance Systems, Lexington MA}}
\affil{MIT Lincoln Laboratory, Lexington, MA, 01801, USA}

\begin{document}

\maketitle

% \section*{Nomenclature}

% \noindent(Nomenclature entries should have the units identified) \todo{fill in}

% {\renewcommand\arraystretch{1.0}
% \noindent\begin{longtable*}{@{}l @{\quad=\quad} l@{}}
% $A$  & amplitude of oscillation \\
% \multicolumn{2}{@{}l}{Subscripts}\\
% cg & center of gravity\\
% \end{longtable*}}

\begin{abstract}
Aircraft collision avoidance systems have long been a key factor in keeping our airspace safe. Over the past decade, the FAA has supported the development of a new family of collision avoidance systems called the Airborne Collision Avoidance System X (ACAS X), which model the collision avoidance problem as a Markov decision process (MDP). Variants of ACAS X have been created for both manned (ACAS Xa) and unmanned aircraft (ACAS Xu and ACAS sXu). The variants primarily differ in the types of collision avoidance maneuvers they issue. For example, ACAS Xa issues vertical collision avoidance advisories, while ACAS Xu and ACAS sXu allow for horizontal advisories due to reduced aircraft performance capabilities. Currently, a new variant of ACAS X, called ACAS Xr, is being developed to provide collision avoidance capability to rotorcraft and Advanced Air Mobility (AAM) vehicles. Due to the desire to minimize deviation from the prescribed flight path of these aircraft, speed adjustments have been proposed as a potential collision avoidance maneuver for aircraft using ACAS Xr. In this work, we investigate the effect of speed change advisories on the safety and operational efficiency of collision avoidance systems. We develop an MDP-based collision avoidance logic that issues speed advisories and compare its performance to that of horizontal and vertical logics through Monte Carlo simulation on existing airspace encounter models. Our results show that while speed advisories are able to reduce collision risk, they are neither as safe nor as efficient as their horizontal and vertical counterparts.
\end{abstract}

\blfootnote{DISTRIBUTION STATEMENT A. Approved for public release. Distribution is unlimited. This material is based upon work supported by the Federal Aviation Administration under Air Force Contract No. FA8702-15-D-0001. Any opinions, findings, conclusions or recommendations expressed in this material are those of the author(s) and do not necessarily reflect the views of the Federal Aviation Administration. This document is derived from work done for the FAA (and possibly others), it is not the direct product of work done for the FAA. The information provided herein may include content supplied by third parties. Although the data and information contained herein has been produced or processed from sources believed to be reliable, the Federal Aviation Administration makes no warranty, expressed or implied, regarding the accuracy, adequacy, completeness, legality, reliability, or usefulness of any information, conclusions or recommendations provided herein. Distribution of the information contained herein does not constitute an endorsement or warranty of the data or information provided herein by the Federal Aviation Administration or the U.S. Department of Transportation. Neither the Federal Aviation Administration nor the U.S. Department of Transportation shall be held liable for any improper or incorrect use of the information contained herein and assumes no responsibility for anyone’s use of the information. The Federal Aviation Administration and U.S. Department of Transportation shall not be liable for any claim for any loss, harm, or other damages arising from access to or use of data information, including without limitation any direct, indirect, incidental, exemplary, special or consequential damages, even if advised of the possibility of such damages. The Federal Aviation Administration shall not be liable for any decision made or action taken, in reliance on the information contained herein. 978-1-5386-6854-2/19/S31.00 2019IEEE}

\pagebreak

\section{Introduction}

\lettrine{C}{ollision} avoidance systems are a fundamental component of the multi-layered conflict management system that ensures safe operations in the National Airspace System (NAS). The first mandated collision avoidance system, the Traffic Alert and Collision Avoidance System (TCAS), was introduced in the 1980s and provided pilots with vertical maneuvers to avoid mid-air collisions \cite{braysy2005vehicle}. In the early 2000s, due to the increased density and complexity of operations, the FAA sponsored the development of the Airborne Collision Avoidance System X (ACAS X). ACAS X is a family of collision avoidance systems that model the collision avoidance problem as a Markov decision process (MDP). 

Each variant of \ACAS has tackled new constraints emerging from aircraft dynamic capabilities, well-clear requirements, and human-in-the-loop considerations. \ACASA\!, for example, was designed to provide vertical alerts equivalent to \TCAS to minimize pilot retraining and confusion when operating commercial flights while reducing nuisance alerts and collision risk compared to \TCAS \cite{Kochenderfer2013}. Reduced aircraft performance capabilities and the adoption of ADS-B led \ACASU and \ACASSU to issue horizontal maneuvers to avoid mid-air collisions \cite{owen2019acas} \cite{alvarez2019acas}. These unmanned variants were designed as detect and avoid (DAA) systems that use sensors to detect other aircraft and provide situational awareness, allowing a remote pilot or automated response to see and avoid as mandated in Part 91 of Title 14 of the US Code of Federal Regulations. 

The newest variant of \ACAS under development, \ACASR\!, is slated to add DAA capability to manned rotorcraft and manned or unmanned Advanced Air Mobility (AAM) vehicles, which includes Urban Air Mobility (UAM) vehicles. In the UAM concept, operations require traversing a high density environment, sometimes referred to as the UAM Operating Environment (UOE) \cite{NASAConops}, in which early implementations of UAM vehicles would be required to fly through static corridor structures \cite{UAMConopsFAA}. Prior to flight, the aircraft would have an approved flight trajectory to reach its final destination with emergency procedures defined at various states in the trajectory. Due to this highly structured environment, stakeholders view spatial divergence from their approved flight trajectory as unsatisfactory. Since the vertical and horizontal maneuvers used by \ACASA and \ACASU result in deviations from the aircraft's nominal trajectory, their use in the context may be limited. For this reason along with the fact that AAM aircraft are capable of hovering, the concept of adding speed change advisories along the prescribed flight path as a collision avoidance solution has been proposed by the community. In this work, we investigate the impact of allowing speed change advisories on the safety and operational efficiency of collision avoidance systems.

%\todo{should we say something about how rotorcraft can more easily handle speed changes? if this is even true?}

The rest of the paper is organized as follows. \Cref{sec:background} discusses background on the techniques used to design and assess aircraft collision avoidance systems. \Cref{sec:prob_form} outlines the formulation of a speed collision avoidance logic. \Cref{sec:results} discusses results from using speed advisories in a diverse set of encounters. \Cref{sec:conclude} provides a summary of the challenges of using a speed logic and a recommendation to the community.

\section{Background}\label{sec:background}
Developing a collision avoidance system requires two key components: designing a collision avoidance logic and assessing its performance in realistic scenarios. This section outlines how these problems have been approached for previous collision avoidance systems in the ACAS X family.

\subsection{MDP-Based Collision Avoidance Logic}\label{sec:mdp_cas}
The collision avoidance logic for all variants of ACAS X was generated by modeling the collision avoidance problem as an MDP. An MDP is a way of encoding a sequential decision making problem in which an agent's action at each time step depends only on its current state \cite{DMU}. An MDP is defined by the tuple $(S, A, T, R, \gamma)$, where $S$ is the state space, $A$ is the action space, $T(s,a,s')$ is the probability of transitioning to state $s'$ given that we are in state $s$ and take action $a$, $R(s,a)$ is the reward for taking action $a$ in state $s$, and $\gamma$ is the discount factor. In the context of aircraft collision avoidance, the state typically represents the speeds and relative position of the ownship and intruder aircraft, and the actions are the collision avoidance maneuvers. The transition model is selected to model the stochastic evolution of an encounter scenario, and the reward model consists of a large penalty for being in a near mid-air collision (NMAC) state, defined as loss of aircraft separation to less than \SI{500}{ft} horizontally and \SI{100}{ft} vertically, and smaller penalties that encourage operational efficiency.

A policy for an MDP maps from states to actions. The policy can be compactly represented by defining a state-action value function, $Q(s,a)$, which represents the expected return when taking action $a$ from state $s$. With a discrete state and action space, the optimal value for each state and action, $Q^\ast(s,a)$, can be obtained using a form of dynamic programming called value iteration. The algorithm relies on iterative updates of $Q^\ast(s,a)$ using the Bellman equation \cite{bellman1954theory}:
\begin{equation}
    Q^\ast(s,a) = R(s,a) + \gamma \sum_{s' \in S} T(s,a,s') \max_{a' \in A} Q^\ast(s',a')
\end{equation}
A deterministic policy $\pi$ can be defined by simply choosing the action with the maximum value at state $s$:
\begin{equation}\label{eq:detpol}
    \pi(s) = \argmax_{a \in A} Q^\ast(s,a)
\end{equation}

For \ACAS, the state variables are discretized, and the collision avoidance MDP is solved offline using value iteration. While \ACAS systems are formulated as MDPs for the solving process, the inherent uncertainty in the state measurements turns the problem into a partially observable Markov decision process (POMDP). The state uncertainty is taken into account during real-time execution using the QMDP method \cite{DMU}.
A table of the final optimal values $Q^\ast(s,a)$ is loaded onto the aircraft, and the policy is used to provide optimal collision avoidance advisories during flight. 
%The lookup table will therefore have a total of $|S||A|$ entries.

\subsection{Encounter Sets} \label{sec:encsets}
Both TCAS and ACAS X have been assessed through Monte Carlo analysis using airspace simulations \cite{TCASMonteCarlo,ACASXMonteCarlo}. A key driver behind these simulations are airspace encounter models, which provide a probabilistic representation of typical aircraft behavior during a close encounter with another aircraft. A broad collection of encounter sets are used to assess the performance of collision avoidance systems. Evaluation with a broad set of encounters not only allows for robust development but also highlights particular encounter characteristics that are challenging for a specific system. The geometries and flight phases explored by each encounter set during the development of the speed logic are summarized in \cref{tab:encounterSets}. These encounter sets have been previously used in the evaluation of other \ACAS variants and are generally accepted as standards to measure the performance of collision avoidance systems \cite{kochenderfer2008correlated}, \cite{weinertRepresentativeSmallUAS2020}, \cite{Katz2019UAM}.

\begin{table}[h]
    \centering
    \caption{Evaluated encounter sets and their general geometries and flight phases. \label{tab:encounterSets}}
    \begin{tabular}{@{}llrr@{}}
          \toprule
          \textbf{Encounter Set Name} & \textbf{Phase(s) of Flight} & \textbf{Ownship Type} & \textbf{Intruder Type}\\
          \midrule
          LLCEM HMD Ext 150 & Correlated Flights in Cruising Phase & Fixed Wing & Fixed Wing\\
          %Rotorcraft & Uncorrelated Takeoff/Landing/Cruising & Rotorcraft or UAM & Rotorcraft or UAM or sUAS \\
          OpSuit & Uncorrelated Cruising Phase & Rotorcraft & Rotorcraft \\
          Hovering & Uncorrelated Hovering vs. Cruising & Rotorcraft & Fixed Wing \\
          \bottomrule
    \end{tabular}
\end{table}

The LLCEM encounter set was sampled from a statistical airspace model created from nine months of radar data covering much of the continental United States and captures the correlation between aircraft trajectories due to intervention from air traffic control \cite{kochenderfer2008correlated}. The OpSuit (operational suitability) encounter set is composed of ownship tracks from \citeauthor{weinertRepresentativeSmallUAS2020} \cite{weinertRepresentativeSmallUAS2020} paired with intruder tracks from \citeauthor{kochenderfer2008correlated} \cite{kochenderfer2008correlated}. The encounters are sampled to have a uniform distribution of horizontal miss distances between \num{0} and \SI{10000}{ft} and vertical miss distances between \num{0} and \SI{2000}{ft}. The hovering encounter set was designed to test collision avoidance systems' ability to resolve NMACs between a hovering ownship and an intruder who may be loitering, transiting, landing, or taking off. Because the ownship has no inherent direction of travel, we initalize it with a north heading so that speed changes result in translations north. To create a diverse set of hovering encounter geometries, the relative heading angle of the intruder at closest point of approach is uniformly distributed between $0$ and $360$ degrees. Overall, these encounter sets allow for a statistical evaluation of collision avoidance systems in terms of safety and operational suitability (e.g., alerting rate per encounter or alerting efficiency).

\section{Problem Formulation}\label{sec:prob_form}
To create an MDP-based collision avoidance logic that issues speed change advisories, we must carefully define each component of the MDP tuple presented in \cref{sec:mdp_cas}. Because we define speed advisories as changes to the aircraft's horizontal speed, we use the same state variables that \ACASU and \ACASSU use for their horizontal logic \cite{owen2019acas, alvarez2019acas}. \Cref{fig:speed_state} shows a visual representation of some of the state variables, while \cref{tab:speed_ss} summarizes the variables and their discretizations.
\begin{figure}[htb]
	\centering
	\begin{tikzpicture}
	\node [UAM top,fill=black,draw=black,minimum width=1cm,rotate=270,scale=0.5] at (0cm,0cm) {};
	\node [UAM top,fill=black,draw=black,minimum width=1cm,rotate=45,scale=0.5] at (5cm,2cm) {}; 
	\draw (0cm,0cm) -- (5cm,2cm);
	\draw [<->] (0cm,0cm) -- (1.5cm,0cm);
	\draw [<->] (5cm,2cm) -- (3.9393cm,3.06066cm);
	\draw (0.8cm,0.0cm) arc (0.0:21.8:0.8cm);
	\draw [dashed] (5cm,2cm) -- (6.5cm,2cm);
	\draw (5.8cm,2cm) arc (0.0:135.0:0.8cm);
	\node [right] at (1.5,0.0) {$v_0$};
	\node [above] at (3.9393cm,3.06066cm) {$v_1$};
	\node [below] at (2.5cm,1cm) {$r$};
	\node [below] at (0,-0.3) {ownship};
	\node [below] at (5,1.7) {intruder};
	\node at (1.1cm,0.2cm) {$\theta$};
	\node at (5.2cm,3.1cm) {$\psi$};
	\end{tikzpicture}
	\caption{Speed logic state. \label{fig:speed_state}}
\end{figure}
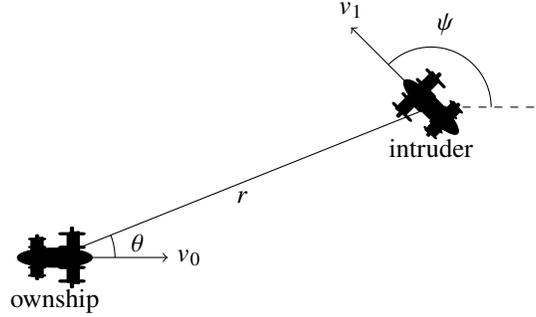
\begin{table}[htb]
    \centering
    \caption{Speed state space. \label{tab:speed_ss}}
    \begin{tabular}{@{}llrrr@{}}
         \toprule
         \textbf{Variable} & \textbf{Description} & \textbf{Units} & \textbf{Range} (low:high) & \textbf{Number of Values} \\
         \midrule
         $r$ & range & ft & 499:48169 & 71 \\
         $\theta$ & bearing & radians & $-\pi:\pi$ & 121 \\
         $\psi$ & relative heading & radians & $-\pi:\pi$ & 121 \\
         $v_0$ & ownship horizontal speed & ft/s & 50:237 & 94 \\
         $v_1$ & intruder horizontal speed & ft/s & 0:237 & 4 \\
         $a_{\text{prev}}$ & previous action & N/A & N/A & 4 \\
         $\tau$ & time to vertical NMAC & s & 0:100 & 10 \\
         \bottomrule
    \end{tabular}
\end{table}
The first three state variables capture the relative position and orientation of the aircraft, and the next two state variables are their horizontal speeds. Furthermore, the state contains the previous action so that we can penalize changes in action such as strengthenings and reversals in the reward model. Finally, we use a summary variable for the vertical state called $\tau$ that represents the time to vertical NMAC. If the logic includes a maintain speed action it will contain four previous actions otherwise three actions present in the state space. We selected the ranges and discretizations for each state variable based on the operational characteristics of UAM vehicles and the requirements of the speed logic \cite{prakasha2021urban}, \cite{rohitMarketStudy}.  Because the actions result in changes in horizontal speed, the discretization for ownship speed must be fine enough to observe the effect of taking a particular action; however, a discretization that is too fine will result in an intractable problem. To address this, we use a finer discretization for ownship speed than \ACASU and \ACASSU and increase the time step (the value by which $\tau$ is incremented) from $1$ second to $10$ seconds %\todo{I noticed the second half of this sentence was removed - I put it back for now unless it was removed intentionally}.

The actions for the MDP are the speed change advisories. The advisories command horizontal accelerations along the flight path and are summarized in \cref{tab:speed_as}. The magnitude of the deceleration and acceleration action were chosen based on passenger comfort and aircraft capabilities as informed by aircraft manufacturers and pilots that participate in RTCA SC-147.  
\begin{table}[htb]
    \centering
    \caption{Speed action space.  \label{tab:speed_as}}
    \begin{tabular}{@{}llr@{}}
         \toprule
         \textbf{Action} & \textbf{Description} & \textbf{Acceleration} \\
         \midrule
         \textsc{COC} & clear of conflict & - \\
         \textsc{SD} & decelerate & $-0.0625g$ \\
         \textsc{SA} & accelerate & $0.0625g$ \\
         \textsc{MA} & maintain acceleration & - \\
         \bottomrule
    \end{tabular}
\end{table}
% \todo{Luis update and write about maintain advisory}
% \todo{did the transition model match ACAS Xu? like with possible turns?}
%\todo{Luis, update language about the transition model and how it different than the Xu model}
% The transition model used is similar to that used for the vertical and horizontal logics of \ACASU \cite{owen2019acas}.

The transition model uses a zero-mean Gaussian noise model to represent the likelihood of a state changing from a given action. A zero-mean Gaussian noise model is applied to the ownship speed (standard deviation of \SI{1.64}{ft/s}), intruder speed (standard deviation of \SI{3.64}{ft/s}), ownship heading (standard deviation of \SI{1.0027}{ft/s}), and intruder heading (standard deviation of \SI{1.0027}{ft/s}). When an action other than \textsc{COC} is selected, the noise model is ignored to represent deterministic transitions in this dimension. For this reason, a maintain action is observed differently than a \textsc{COC} action. We use the same reward model structure as \ACASU\!\!, which contains a large penalty for NMACs and smaller penalties for undesirable operational characteristics such as issuing an advisory or reversing an advisory \cite{owen2019acas}. The reward model can be tuned to balance between safety and operational efficiency. 
%Although the \textsc{COC} and Maintain action have the same effect, the \textsc{COC} action applies the gaussian transition model 

% \subsection{State Space}
% % action needs to affect state

% \subsection{Action Space}
% % why action magnitude

% \subsection{Transition Model}
% % just reference ACAS Xu paper

% \subsection{Reward Model}
% % just reference ACAS Xu paper

\section{Results}\label{sec:results}
In this section, we present the general performance of the speed logic when used both on its own and in conjunction with horizontal and vertical logics. We analyze the results by comparing its performance to other \ACAS logics and identifying general encounter characteristics that are uniquely challenging to a speed logic.

\subsection{Overall Performance}
\begin{figure}[h!]
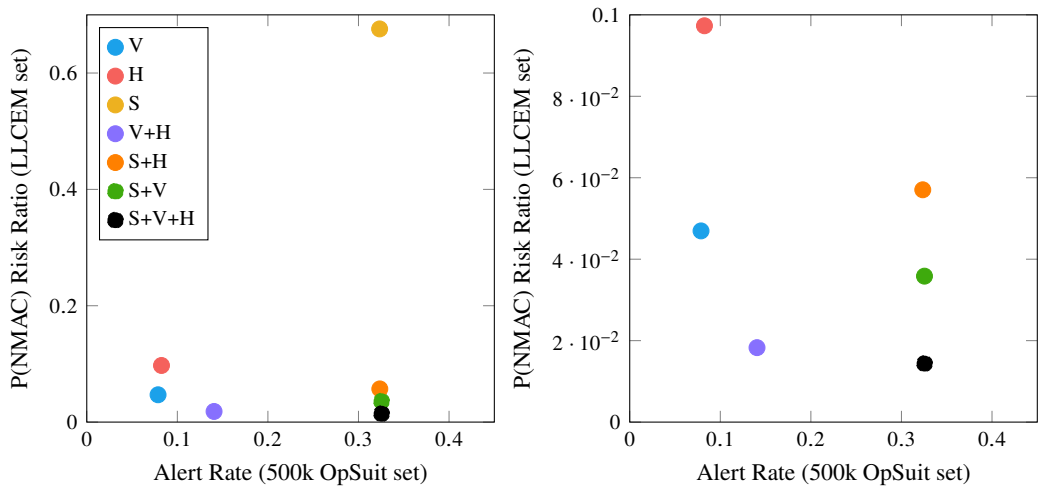

    \centering
    \include{figures/eu}
    \caption{Equipped-Unequipped risk ratio versus alert rate for all logics whether working in conjunction or alone. The left figure includes the speed-only results, while the right figure zooms in on the results in the absence of speed-only. \label{fig:euPerformance}}
\end{figure}

\begin{figure}[h!]
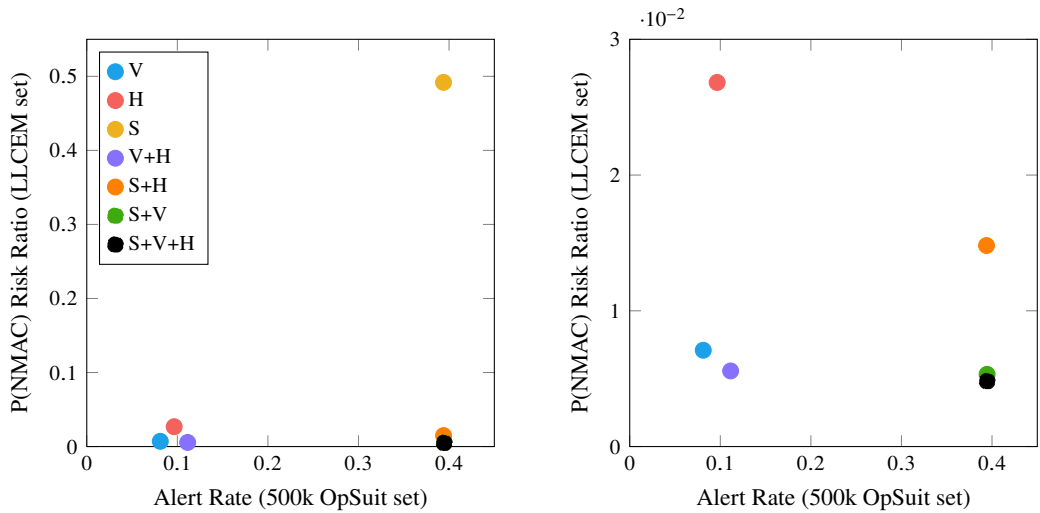

    \centering
    \include{figures/ee}
    \caption{Equipped-Equipped risk ratio versus alert rate for all logics whether working in conjunction or alone. The left figure includes speed-only results, while the right figure zooms in on the results in the absence of speed-only. \label{fig:eePerformance}}
\end{figure}

We explored the performance of the speed logic by itself and in combination with other \ACAS logics in encounters in which both aircraft involved have a collision avoidance system (CAS) onboard (Equipped-Equipped) and encounters in which only one aircraft is equipped (Equipped-Unequipped). \Cref{fig:euPerformance} and \cref{fig:eePerformance} compare the safety and operational efficiency of each logic. We use a metric called the P(NMAC) risk ratio to measure safety of each collision avoidance system on a given encounter set, which is defined as
% \begin{equation}
%     \text{risk ratio} = \frac{\text{number of NMACs with collision avoidance system}}{\text{number of NMACs without collision avoidance system}}
% \end{equation}
% \begin{equation}
%     P(\text{NMAC}) = \frac{\sum_{i=1}^{N} f_{\text{NMAC}}(HMD,VMD,\text{Equipage}) \cdot w_{i}}{\sum_{i=1}^{N}w_{i}}
% \label{eq:pnmac_calculation_bias}
% \end{equation}

\begin{equation}
    \text{P(NMAC) Risk Ratio} = \frac{\sum_{i=1}^{n} \text{P(NMAC} \mid \text{CAS, altimeter error model})_i \cdot w_i}{\sum_{i=1}^{n} \text{P(NMAC} \mid \text{no-CAS, altimeter error model})_i \cdot w_i}.
    \label{eq:pnmac_risk_ratio}
\end{equation}
where $w_i$ is the weight associated with the specific encounter provided by the encounter model \cite{kochenderfer2008correlated}, the altimeter error model is that described by Section 4.4.2.4 in \cite{annex10volume}, $i$ is the index of the encounter in the encounter set, and $n$ is the total number of encounters. A lower risk ratio indicates a safer system. We assess operational efficiency by calculating the alert rate on the OpSuit encounter set. Overall, the speed logic by itself has risk ratios \num{6} to \num{69} times higher in the LLCEM encounter set and alerts \num{3} to \num{4} times more in the OpSuit encounter set than the horizontal and vertical logic. 

Regardless of equipage type, the speed logic alone does not meet the target risk ratios set out by the International Civil Aviation Organization (ICAO) \cite{annex10volume}, which states that TCAS equivalent systems must meet a risk ratio of 0.18 against unequipped intruders and 0.04 for equipped intruders. Furthermore, its risk ratios are above the ASTM target risk ratios of 0.18 for sUAS operations against cooperative intruders \cite{astm2020}. When the speed logic is used in conjunction with another logic (horizontal or vertical), the resulting blended logic has a slightly lower risk ratio than the individual logics on their own in all cases at the cost of a significant increase in alerts.

\begin{figure}[h!]
    \centering
    \include{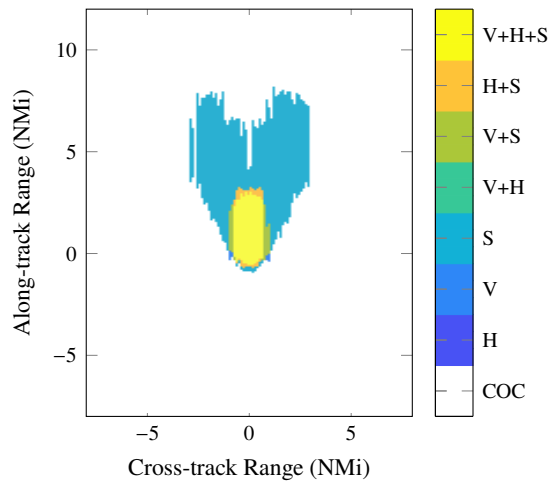}
    \caption{Alerting profiles constructed from simulated co-altitude encounters which include surveillance noise and an ownship aircraft that is unresponsive to alerts. Colored regions depict the extent of each logic's alerting regions. \label{fig:topViewPolicyPlot}}
\end{figure}

%Before assessing the performance of the speed logic on the encounter sets described in \cref{sec:encsets}, we analyzed the alerting profile of the resulting speed policy. 
By simulating co-altitude encounters with various relative headings in which the ownship aircraft is unresponsive, we can understand when the each type of collision avoidance logic will issue an alert. \Cref{fig:topViewPolicyPlot} shows the results of this analysis for each collision avoidance logic. With a speed change magnitude of $0.0625g$, the logic requires long lead times to impact aircraft separation at closest point of approach. For this reason, the speed logic tends to alert at higher ranges than the horizontal and vertical logics. This result is consistent with the high alert rates associated with the 
speed logics shown in \cref{fig:euPerformance} and \cref{fig:eePerformance}.

\subsection{Challenging Encounter Characteristics for a Speed Logic}
Unlike the vertical and horizontal logics of ACAS X, the speed logic keeps the aircraft on its prescribed flight trajectory (e.g., UAM flight corridors, VFR corridor, etc.) While this feature may be desirable from an operational standpoint, it can impact safety in certain encounter geometries. In particular, the speed logic's lack of directional guidance results in an inability to resolve head-on, co-altitude encounter geometries regardless of closure rate. In a head-on situation, the speed logic can only delay an impending collision by reducing the speed of the aircraft but cannot prevent one entirely.
%or increase the speed of the aircraft to reduce the accrued penalty from maintaining a collision course.
Another problematic geometry is a co-altitude overtake where the aircraft is either traveling faster than the intruder and cannot slow down in time or can not fly fast enough to avoid a collision with an intruder (surpassing dynamic capabilities of aircraft). 
% This property of the speed logic is especially apparent in the hovering encounter set.

This characteristic of the speed logic is highlighted by the simulation results on the hovering encounter set. \Cref{tab:HoveringResults} compares the risk ratios of each logic in the hovering encounter set. We note that for a horizontal RA to take effect, the aircraft was required to accelerate to a minimum velocity of $30$ knots at an equivalent acceleration to the speed logic ($0.0625g$).
\begin{table}[h]
    \centering
    \caption{Hovering encounter set performance \label{tab:HoveringResults}}
    \begin{tabular}{@{}lr@{}}
         \toprule
         \textbf{Active Logics} & \textbf{Equipped-Unequipped Risk Ratio}\\
         \midrule
          \textbf{Speed}   & 0.209 \\
          \textbf{Vertical}   & 0.028 \\
          \textbf{Horizontal}   & 0.172 \\
          \bottomrule
    \end{tabular}
\end{table}

\begin{figure}[h!]
    \centering
    \include{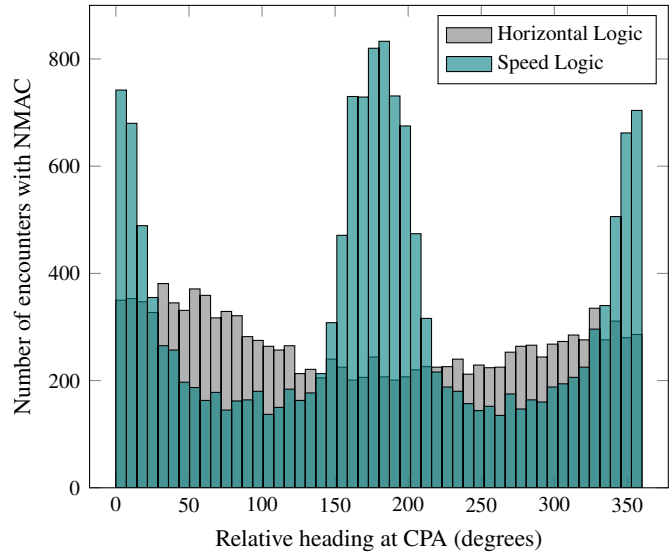}
    \caption{Distribution of NMACs for speed and horizontal logics as a function of relative heading at closest point of approach.} % \todo{Luis to update image to make it overall prettier}{}
    \label{fig:hoveringRelHeading}
\end{figure}

The vertical logic significantly outperforms the horizontal and speed logics in terms of risk ratio. The results appear to indicate that a speed logic and horizontal logic are roughly equivalent in performance in hovering encounters. However, when we consider the relative heading of the aircraft at closest point of approach, it is evident that the horizontal logic has a more uniform performance regardless of geometry. \Cref{fig:hoveringRelHeading} shows the distribution of the relative headings of encounters that resulted in NMACs for the vertical and horizontal logics. The high concentration of NMACs at relative headings close to  $0$\textsc{$^\circ$}, $180$\textsc{$^\circ$}, and $360$\textsc{$^\circ$} indicates that the speed logic struggles to resolve encounters that develop into an overtake or head-on geometry.

\subsection{Effect of Complexity on Performance of CAS Equipped Piloted Aircraft}
In a crewed aircraft, a pilot may not be able to respond to all three advisory dimensions (vertical, horizontal, and speed) at the same time. Therefore, presenting an increased complexity in resolution advisories by having more than one dimension active at once can reduce the likelihood of pilot response to any single dimension. 
%Therefore, there is a non-zero probability of a pilot ignoring a single dimension.
Ultimately, the realized performance gains of introducing a speed advisory to a system that already has vertical and horizontal advisories is a function of both the logic outputs and the pilot choice of response.

By modeling the pilot's decision to respond to an advisory as independently probabilistic in each dimension, we can assess the benefit of adding a speed advisory in this context. To model this effect, we first select a probability that the pilot does not respond in any dimension. Using this probability, we can calculate the probability of only responding to each possible subset of advisory dimensions, assuming the pilot has the same probability of not responding to each advisory. We calculate the probability of response for each possible subset of advisory dimensions as follows
\begin{equation}
    \begin{split}
        P_{\text{response}} &= 1-(P_{\text{no-response}})^{1/N_{\text{actions}}} \\
        P_{\text{single-dimension-response}} &= (1-P_{\text{response}})^{2}P_{\text{response}} \\
        P_{\text{two-dimension-response}} &= (1-P_{\text{response}})P_{\text{response}}^2 \\
        P_{\text{three-dimension-response}} &= P_{\text{response}}^3
    \end{split}
\end{equation}
An example of the resulting probabilities using this technique is shown in the third and fourth column of \cref{tab:prob_pr_response}. We can calculate the probability of NMAC according to this model as a weighted combination of the simulated probability of NMAC for each possible combination of logics. The bottom row of \cref{tab:prob_pr_response} shows the results. Overall, the introduction of a third response dimension would increase the overall P(NMAC) of the system by 44\% from \num{3.31e-4} to \num{4.77e-4}. 

% \begin{equation}
%     \label{eq:Prob}
%     P_{response} = 1-(P_{no-response})^{1/N_{actions}}
% \end{equation}

% \begin{equation}
%     \label{eq:oneDimensionProb}
%     P_{single-dimension-response} = (1-P_{response})^{2}P_{response}
% \end{equation}

% \begin{equation}
%     \label{eq:twoDimensionProb}
%     P_{two-dimension-response} = (1-P_{response})P_{response}^2
% \end{equation}

% \begin{equation}
%     \label{eq:threeDimensionProb}
%     P_{three-dimension-response} = P_{response}^3
% \end{equation}
%First, the probability of not responding to any dimension is chosen and each probability of response for all dimensions are calculated independently, as shown in \Cref{tab:prob_pr_response}. Overall, \Cref{tab:prob_pr_response} shows the introduction of a third response dimension would increase the overall P(NMAC) of the system by 44\% from \num{3.31e-4} to \num{4.77e-4}. 

\begin{table}[htb]
    \centering
    \caption{Probabilistic pilot response results. \label{tab:prob_pr_response}}
    \begin{tabular}{@{}lrrr@{}}
         \toprule
         \textbf{Response} & \textbf{P(NMAC)} & \textbf{Probability with Speed} & \textbf{Probability without Speed} \\
         \midrule
         \textsc{None} & \num{3.01e-3} & 0.10 & 0.10 \\
         \textsc{H} & \num{8.08e-5} & 0.12  & 0.22 \\
         \textsc{V} & \num{2.14e-5} & 0.12  & 0.22 \\
         \textsc{S} & \num{1.32e-3} & 0.12  & - \\
         \textsc{H+S} & \num{4.33e-5} & 0.13  & - \\
         \textsc{V+S} & \num{1.50e-5} & 0.13  & - \\
         \textsc{H+V} & \num{1.68e-5} & 0.13  & 0.46 \\
         \textsc{H+V+S} & \num{1.41e-5} & 0.15  & - \\ \midrule
         \textsc{Total System P(NMAC)} & - & \num{4.77e-4}  & \num{3.31e-4} \\ 
         \bottomrule
    \end{tabular}
\end{table}

\begin{figure}[h!]
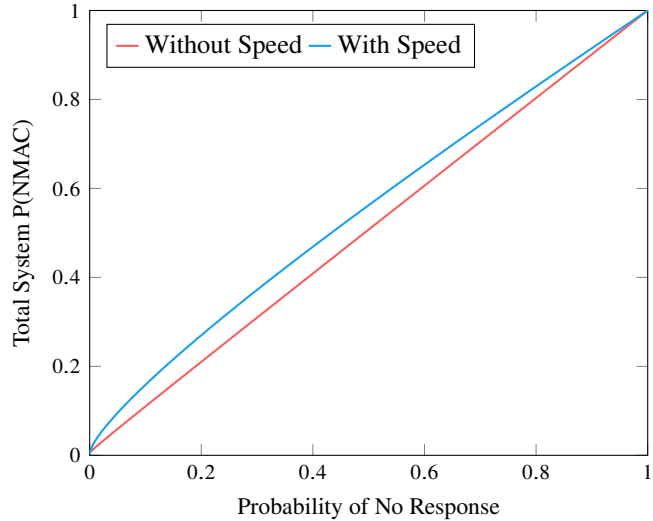

    \centering
    \include{figures/pr}
    \caption{Total system P(NMAC) as a function of probability of pilot not responding to an advisory. \label{fig:PResponseModel}}
\end{figure}

By varying the probability of no advisory response and using the method previously presented, we can calculate the weighted risk ratio for various pilot response assumptions. As seen in \cref{fig:PResponseModel}, increasing the probability of no response to an advisory can increase the difference in safety between a system with vertical and horizontal advisories and a system that introduces speed advisories as well. This is due to the possibility of a pilot choosing any combination of horizontal or vertical with speed advisories or choosing speed advisories alone instead of vertical or horizontal.

\section{Conclusion}\label{sec:conclude}
Speed adjustments allow aircraft to perform collision avoidance maneuvers without deviating from their prescribed flight path. For this reason, they have been proposed as a potential advisory for ACAS Xr, which is being developed to provide DAA capability to rotorcraft and Advanced Air Mobility (AAM) aircraft. In this work, we investigated the impact of using speed adjustments as a collision avoidance maneuver in MDP-based collision avoidance logics. 

From the results presented, we conclude that a speed collision avoidance logic can indeed reduce the risk of collision in a broad set of encounters. However, the speed logic on its own has significantly higher risk ratios than the horizontal and vertical logics.  Furthermore, because speed advisories require long lead times to have an impact, the speed performance comes at the cost of an increased alert rate. Moreover, we showed that a speed logic on its own cannot resolve co-altitude head-on and overtake encounter geometries. Finally, we showed that when the speed logic is used in conjunction with other logics, the blended system is slightly safer than each individual logic on its own but alerts significantly more frequently. However, when taking into account the probability of a pilot not responding to an alerting dimension, the introduction of the speed logic can reduce the overall safety of the system. 

In general, for a system where response to a speed advisory would not reduce the probability of response to other advisories, such as a system with an automated response, speed advisories could provide a safety benefit. In contrast, for systems where non-response to some advisories is a likely outcome, such as with a human pilot, the benefits of speed advisories may not be worth the operational cost.

\section*{Acknowledgments}

The authors wish to thank Neal Suchy for the support of ACAS X development and in particular the speed logic presented in this work. The authors also wish to thank the engineers at Airbus A3 for their initial inspiration and support of this work.

\bibliography{refs}

\end{document}

%% file: figures/eu.tex
\definecolor{mycolor1}{rgb}{0.00000,0.44700,0.74100}%
\definecolor{mycolor2}{rgb}{0.85000,0.32500,0.09800}%
\definecolor{mycolor3}{rgb}{0.92900,0.69400,0.12500}%
\definecolor{mycolor4}{rgb}{0.49400,0.18400,0.55600}%
\definecolor{mycolor5}{rgb}{0.46600,0.67400,0.18800}%
\definecolor{mycolor6}{rgb}{0.30100,0.74500,0.93300}%
\definecolor{mycolor7}{rgb}{0.63500,0.07800,0.18400}%
\begin{tikzpicture}

\begin{groupplot}[group style={horizontal sep = 1.8cm, group size=2 by 1}]

\nextgroupplot [%
    height = {7cm},
    width = {7cm},
    xmin = 0.0,
    xmax = 0.45,
    xlabel={Alert Rate (500k OpSuit set)},
    ymin = 0,
    ymax = 0.7,
    ylabel={P(NMAC) Risk Ratio (LLCEM set)},
    legend style={font=\footnotesize},
    legend pos = {north west}
]

\addplot+[only marks, mark=*, mark options={rotate=180}, mark size = 3, draw=pastelBlue, fill=pastelBlue] table[row sep=crcr]{%
x	y\\
0.078722	0.0469474173826481\\
};

\addplot+[only marks, mark=*, mark options={}, mark size = 3, draw=pastelRed, fill=pastelRed] table[row sep=crcr]{%
x	y\\
0.082506	0.0973151711277574\\
};

\addplot+[only marks, mark=*, mark options={}, mark size=3, draw=mycolor3, fill=mycolor3] table[row sep=crcr]{%
x	y\\
0.323348	0.675826939157848\\
};

\addplot+[only marks, mark=*, mark options={}, mark size=3, draw=pastelPurple, fill=pastelPurple] table[row sep=crcr]{%
x	y\\
0.140542	0.018280679041001\\
};

\addplot+[only marks, mark=*, mark options={}, mark size=3, draw=orange, fill=orange] table[row sep=crcr]{%
x	y\\
0.32355	0.0570421127604014\\
};

\addplot+[only marks, mark=*, mark options={}, mark size=3, draw=pastelGreen, fill=pastelGreen] table[row sep=crcr]{%
x	y\\
0.325442	0.0358369638985304\\
};

\addplot+[only marks, mark=*, mark options={}, mark size=3, draw=black, fill=black] table[row sep=crcr]{%
x	y\\
0.325526	0.014367643137025\\
};

\legend{{}{V}, {}{H}, {}{S}, {}{V+H}, {}{S+H}, {}{S+V}, {}{S+V+H}}

%\legend{V-nM, H-nM, VH-nM, S-wM, S-nM, SH-wM, SH-nM, SV-wM, SV-nM, SVH-wM, SVH-nM}

\nextgroupplot [%
    height = {7cm},
    width = {7cm},
    xmin = 0.0,
    xmax = 0.45,
    xlabel={Alert Rate (500k OpSuit set)},
    ymin = 0,
    ymax = 0.1,
    ylabel={P(NMAC) Risk Ratio (LLCEM set)},
    legend style={font=\footnotesize},
    legend pos = {north west}
]

\addplot+[only marks, mark=*, mark options={rotate=180}, mark size = 3, draw=pastelBlue, fill=pastelBlue] table[row sep=crcr]{%
x	y\\
0.078722	0.0469474173826481\\
};

\addplot+[only marks, mark=*, mark options={}, mark size = 3, draw=pastelRed, fill=pastelRed] table[row sep=crcr]{%
x	y\\
0.082506	0.0973151711277574\\
};

\addplot+[only marks, mark=*, mark options={}, mark size=3, draw=pastelPurple, fill=pastelPurple] table[row sep=crcr]{%
x	y\\
0.140542	0.018280679041001\\
};

\addplot+[only marks, mark=*, mark options={}, mark size=3, draw=orange, fill=orange] table[row sep=crcr]{%
x	y\\
0.32355	0.0570421127604014\\
};

\addplot+[only marks, mark=*, mark options={}, mark size=3, draw=pastelGreen, fill=pastelGreen] table[row sep=crcr]{%
x	y\\
0.325442	0.0358369638985304\\
};

\addplot+[only marks, mark=*, mark options={}, mark size=3, draw=black, fill=black] table[row sep=crcr]{%
x	y\\
0.325526	0.014367643137025\\
};

%\legend{{}{V-nM}, {}{H-nM}, {}{VH-nM}, {}{S-wM}, {}{S-nM}, {}{SH-wM}, {}{SH-nM}, {}{SV-wM}, {}{SV-nM}, {}{SVH-wM}, {}{SVH-nM}}

\end{groupplot}

\end{tikzpicture}%

%% file: figures/ee.tex
\definecolor{mycolor1}{rgb}{0.00000,0.44700,0.74100}%
\definecolor{mycolor2}{rgb}{0.85000,0.32500,0.09800}%
\definecolor{mycolor3}{rgb}{0.92900,0.69400,0.12500}%
\definecolor{mycolor4}{rgb}{0.49400,0.18400,0.55600}%
\definecolor{mycolor5}{rgb}{0.46600,0.67400,0.18800}%
\definecolor{mycolor6}{rgb}{0.30100,0.74500,0.93300}%
\definecolor{mycolor7}{rgb}{0.63500,0.07800,0.18400}%
\begin{tikzpicture}

\begin{groupplot}[group style={horizontal sep = 1.8cm, group size=2 by 1}]

\nextgroupplot [%
    height = {7cm},
    width = {7cm},
    xmin = 0.0,
    xmax = 0.45,
    xlabel={Alert Rate (500k OpSuit set)},
    ymin = 0,
    ymax = 0.55,
    ylabel={P(NMAC) Risk Ratio (LLCEM set)},
    legend style={font=\footnotesize},
    legend pos = {north west}
]

\addplot+[only marks, mark=*, mark options={rotate=180}, mark size = 3, draw=pastelBlue, fill=pastelBlue] table[row sep=crcr]{%
x	y\\
0.081172	0.00709560645884846\\
};

\addplot+[only marks, mark=*, mark options={}, mark size = 3, draw=pastelRed, fill=pastelRed] table[row sep=crcr]{%
x	y\\
0.096512	0.0268213805130793\\
};

\addplot+[only marks, mark=*, mark options={}, mark size=3, draw=mycolor3, fill=mycolor3] table[row sep=crcr]{%
x	y\\
0.39388	0.491850621954135\\
};

\addplot+[only marks, mark=*, mark options={}, mark size=3, draw=pastelPurple, fill=pastelPurple] table[row sep=crcr]{%
x	y\\
0.111402	0.00557046793122701\\
};

\addplot+[only marks, mark=*, mark options={}, mark size=3, draw=orange, fill=orange] table[row sep=crcr]{%
x	y\\
0.39394	0.014810494462142\\
};

\addplot+[only marks, mark=*, mark options={}, mark size=3, draw=pastelGreen, fill=pastelGreen] table[row sep=crcr]{%
x	y\\
0.394558	0.00532041828719054\\
};

\addplot+[only marks, mark=*, mark options={}, mark size=3, draw=black, fill=black] table[row sep=crcr]{%
x	y\\
0.394598	0.004812616256383\\
};

\legend{{}{V}, {}{H}, {}{S}, {}{V+H}, {}{S+H}, {}{S+V}, {}{S+V+H}}

%\legend{V-nM, H-nM, VH-nM, S-wM, S-nM, SH-wM, SH-nM, SV-wM, SV-nM, SVH-wM, SVH-nM}

\nextgroupplot [%
    height = {7cm},
    width = {7cm},
    xmin = 0.0,
    xmax = 0.45,
    xlabel={Alert Rate (500k OpSuit set)},
    ymin = 0,
    ymax = 0.03,
    ylabel={P(NMAC) Risk Ratio (LLCEM set)},
    legend style={font=\footnotesize},
    legend pos = {north west}
]

\addplot+[only marks, mark=*, mark options={rotate=180}, mark size = 3, draw=pastelBlue, fill=pastelBlue] table[row sep=crcr]{%
x	y\\
0.081172	0.00709560645884846\\
};

\addplot+[only marks, mark=*, mark options={}, mark size = 3, draw=pastelRed, fill=pastelRed] table[row sep=crcr]{%
x	y\\
0.096512	0.0268213805130793\\
};

\addplot+[only marks, mark=*, mark options={}, mark size=3, draw=pastelPurple, fill=pastelPurple] table[row sep=crcr]{%
x	y\\
0.111402	0.00557046793122701\\
};

\addplot+[only marks, mark=*, mark options={}, mark size=3, draw=orange, fill=orange] table[row sep=crcr]{%
x	y\\
0.39394	0.014810494462142\\
};

\addplot+[only marks, mark=*, mark options={}, mark size=3, draw=pastelGreen, fill=pastelGreen] table[row sep=crcr]{%
x	y\\
0.394558	0.00532041828719054\\
};

\addplot+[only marks, mark=*, mark options={}, mark size=3, draw=black, fill=black] table[row sep=crcr]{%
x	y\\
0.394598	0.004812616256383\\
};

%\legend{{}{V-nM}, {}{H-nM}, {}{VH-nM}, {}{S-wM}, {}{S-nM}, {}{SH-wM}, {}{SH-nM}, {}{SV-wM}, {}{SV-nM}, {}{SVH-wM}, {}{SVH-nM}}

\end{groupplot}

\end{tikzpicture}%

%% file: figures/TopViewCombinedPolicyPlots.tex
\begin{tikzpicture}[]
\begin{axis}[%
    height= {7cm},
    point meta min=0,
    point meta max=7,
    axis equal image,
    xmin= -8, %-5.65219816979163,
    xmax= 8, %5.65219816979163,
    xlabel= {Cross-track Range (NMi)},
    ymin= -8, %-7.80711174090747,
    ymax= 12, %11.6911693302905,
    ylabel= {Along-track Range (NMi)},
    colormap={mymap}{ 
        rgb=(1,1,1); 
        rgb=(0.281014,0.322757,0.957886); 
        rgb=(0.178643,0.528857,0.968157); 
        rgb=(0.0688714,0.694771,0.839357); 
        rgb=(0.216086,0.7843,0.5923); 
        rgb=(0.671986,0.779271,0.2227); 
        rgb=(0.996986,0.765857,0.219943); 
        rgb=(0.9769,0.9839,0.0805)},
    colorbar sampled,
    colormap access=piecewise const,
    colorbar style={
            ytick = {0.4375, 1.3125, 2.1875, 3.0625, 3.9375, 4.8125, 5.6875, 6.5625},
            yticklabels = {COC, H, V, S, V+H, V+S, H+S, V+H+S}
            },
    colorbar,
]

\addplot [
    ] graphics [
    xmin=-5.65219816979163, 
    xmax=5.65219816979163, 
    ymin=11.6911693302905, 
    ymax=-7.80711174090747
] {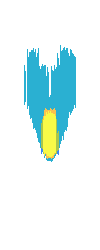};

\end{axis}

% \begin{axis}[%
% width=5.833in,
% height=4.375in,
% at={(0in,0in)},
% scale only axis,
% point meta min=0,
% point meta max=1,
% xmin=0,
% xmax=1,
% ymin=0,
% ymax=1,
% axis line style={draw=none},
% ticks=none,
% axis x line*=bottom,
% axis y line*=left
% ]
% \end{axis}
\end{tikzpicture}%

%% file: figures/hoveringRelHeadingPlots.tex
\begin{tikzpicture}

\begin{axis}[%
    height = {8cm},
    xmin = -18,
    xmax = 378,
    xlabel = {Relative heading at CPA (degrees)},
    ymin = 0,
    ymax = 900,
    ylabel = {Number of encounters with NMAC},
    legend style={font=\footnotesize}
]

\addplot[ybar interval, fill=gray, fill opacity=0.6, area legend] table[row sep=crcr] {%
x	y\\
0	350\\
7.2	353\\
14.4	347\\
21.6	327\\
28.8	381\\
36	345\\
43.2	331\\
50.4	371\\
57.6	359\\
64.8	317\\
72	329\\
79.2	321\\
86.4	282\\
93.6	275\\
100.8	264\\
108	257\\
115.2	265\\
122.4	213\\
129.6	221\\
136.8	205\\
144	240\\
151.2	225\\
158.4	201\\
165.6	206\\
172.8	244\\
180	207\\
187.2	201\\
194.4	207\\
201.6	220\\
208.8	226\\
216	225\\
223.2	226\\
230.4	240\\
237.6	212\\
244.8	229\\
252	224\\
259.2	225\\
266.4	253\\
273.6	264\\
280.8	266\\
288	244\\
295.2	268\\
302.4	273\\
309.6	285\\
316.8	276\\
324	335\\
331.2	276\\
338.4	311\\
345.6	280\\
352.8	286\\
360	286\\
};

\addplot[ybar interval, fill=teal, fill opacity=0.6, area legend] table[row sep=crcr] {%
x	y\\
0	742\\
7.2	680\\
14.4	489\\
21.6	355\\
28.8	265\\
36	257\\
43.2	197\\
50.4	187\\
57.6	163\\
64.8	178\\
72	145\\
79.2	162\\
86.4	164\\
93.6	180\\
100.8	137\\
108	150\\
115.2	184\\
122.4	163\\
129.6	177\\
136.8	213\\
144	308\\
151.2	471\\
158.4	730\\
165.6	729\\
172.8	820\\
180	833\\
187.2	731\\
194.4	675\\
201.6	474\\
208.8	316\\
216	216\\
223.2	188\\
230.4	180\\
237.6	157\\
244.8	144\\
252	152\\
259.2	135\\
266.4	175\\
273.6	147\\
280.8	164\\
288	160\\
295.2	188\\
302.4	194\\
309.6	206\\
316.8	225\\
324	296\\
331.2	340\\
338.4	506\\
345.6	662\\
352.8	704\\
360	704\\
};

\legend{Horizontal Logic, Speed Logic}
\end{axis}

\end{tikzpicture}%

%% file: figures/pr.tex
\begin{tikzpicture}[]
\begin{axis}[
  height = {8.5cm},
  legend pos = {north west},
  ylabel = {Total System P(NMAC)},
  xmin = {0.0},
  xmax = {1.0},
  xlabel = {Probability of No Response},
  ymin = {0.0},
  ymax = {1.0},
  width = {9cm},
  height = {7.5cm},
  legend columns = 3
]

\addplot+[
  mark = {none},
  solid, thick, pastelRed
] coordinates {
  (0     ,    0.0056)
(0.0050,    0.0121)
(0.0100,    0.0176)
(0.0150,    0.0229)
(0.0200,    0.0282)
(0.0250,    0.0335)
(0.0300,    0.0387)
(0.0350,    0.0439)
(0.0400,    0.0490)
(0.0450,    0.0541)
(0.0500,    0.0593)
(0.0550,    0.0644)
(0.0600,    0.0695)
(0.0650,    0.0745)
(0.0700,    0.0796)
(0.0750,    0.0847)
(0.0800,    0.0898)
(0.0850,    0.0948)
(0.0900,    0.0999)
(0.0950,    0.1049)
(0.1000,   0.1100)
(0.1050,   0.1150)
(0.1100,   0.1200)
(0.1150,   0.1250)
(0.1200,   0.1301)
(0.1250,   0.1351)
(0.1300,   0.1401)
(0.1350,   0.1451)
(0.1400,   0.1501)
(0.1450,   0.1551)
(0.1500,   0.1602)
(0.1550,   0.1652)
(0.1600,   0.1702)
(0.1650,   0.1752)
(0.1700,   0.1802)
(0.1750,   0.1852)
(0.1800,   0.1901)
(0.1850,   0.1951)
(0.1900,   0.2001)
(0.1950,   0.2051)
(0.2000,    0.2101)
(0.2050,    0.2151)
(0.2100,    0.2201)
(0.2150,    0.2250)
(0.2200,    0.2300)
(0.2250,    0.2350)
(0.2300,    0.2400)
(0.2350,    0.2450)
(0.2400,    0.2499)
(0.2450,    0.2549)
(0.2500,    0.2599)
(0.2550,    0.2649)
(0.2600,    0.2698)
(0.2650,    0.2748)
(0.2700,    0.2798)
(0.2750,    0.2847)
(0.2800,    0.2897)
(0.2850,    0.2947)
(0.2900,    0.2996)
(0.2950,    0.3046)
(0.3000,    0.3096)
(0.3050,    0.3145)
(0.3100,    0.3195)
(0.3150,    0.3244)
(0.3200,    0.3294)
(0.3250,    0.3344)
(0.3300,    0.3393)
(0.3350,    0.3443)
(0.3400,    0.3492)
(0.3450,    0.3542)
(0.3500,    0.3591)
(0.3550,    0.3641)
(0.3600,    0.3690)
(0.3650,    0.3740)
(0.3700,    0.3789)
(0.3750,    0.3839)
(0.3800,    0.3888)
(0.3850,    0.3938)
(0.3900,    0.3987)
(0.3950,    0.4037)
(0.4000,    0.4086)
(0.4050,    0.4136)
(0.4100,    0.4185)
(0.4150,    0.4235)
(0.4200,    0.4284)
(0.4250,    0.4334)
(0.4300,    0.4383)
(0.4350,    0.4433)
(0.4400,    0.4482)
(0.4450,    0.4532)
(0.4500,    0.4581)
(0.4550,    0.4630)
(0.4600,    0.4680)
(0.4650,    0.4729)
(0.4700,    0.4779)
(0.4750,    0.4828)
(0.4800,    0.4878)
(0.4850,    0.4927)
(0.4900,    0.4976)
(0.4950,    0.5026)
(0.5000,    0.5075)
(0.5050,    0.5124)
(0.5100,    0.5174)
(0.5150,    0.5223)
(0.5200,    0.5273)
(0.5250,    0.5322)
(0.5300,    0.5371)
(0.5350,    0.5421)
(0.5400,    0.5470)
(0.5450,    0.5519)
(0.5500,    0.5569)
(0.5550,    0.5618)
(0.5600,    0.5667)
(0.5650,    0.5717)
(0.5700,    0.5766)
(0.5750,    0.5815)
(0.5800,    0.5865)
(0.5850,    0.5914)
(0.5900,    0.5963)
(0.5950,    0.6013)
(0.6000,    0.6062)
(0.6050,    0.6111)
(0.6100,    0.6161)
(0.6150,    0.6210)
(0.6200,    0.6259)
(0.6250,    0.6309)
(0.6300,    0.6358)
(0.6350,    0.6407)
(0.6400,    0.6457)
(0.6450,    0.6506)
(0.6500,    0.6555)
(0.6550,    0.6604)
(0.6600,    0.6654)
(0.6650,    0.6703)
(0.6700,    0.6752)
(0.6750,    0.6802)
(0.6800,    0.6851)
(0.6850,    0.6900)
(0.6900,    0.6949)
(0.6950,    0.6999)
(0.7000,    0.7048)
(0.7050,    0.7097)
(0.7100,    0.7146)
(0.7150,    0.7196)
(0.7200,    0.7245)
(0.7250,    0.7294)
(0.7300,    0.7343)
(0.7350,    0.7393)
(0.7400,    0.7442)
(0.7450,    0.7491)
(0.7500,    0.7540)
(0.7550,    0.7590)
(0.7600,    0.7639)
(0.7650,    0.7688)
(0.7700,    0.7737)
(0.7750,    0.7787)
(0.7800,    0.7836)
(0.7850,    0.7885)
(0.7900,    0.7934)
(0.7950,    0.7983)
(0.8000,    0.8033)
(0.8050,    0.8082)
(0.8100,    0.8131)
(0.8150,    0.8180)
(0.8200,    0.8230)
(0.8250,    0.8279)
(0.8300,    0.8328)
(0.8350,    0.8377)
(0.8400,    0.8426)
(0.8450,    0.8476)
(0.8500,    0.8525)
(0.8550,    0.8574)
(0.8600,    0.8623)
(0.8650,    0.8672)
(0.8700,    0.8722)
(0.8750,    0.8771)
(0.8800,    0.8820)
(0.8850,    0.8869)
(0.8900,    0.8918)
(0.8950,    0.8967)
(0.9000,   0.9017)
(0.9050,   0.9066)
(0.9100,   0.9115)
(0.9150,   0.9164)
(0.9200,   0.9213)
(0.9250,   0.9263)
(0.9300,   0.9312)
(0.9350,   0.9361)
(0.9400,   0.9410)
(0.9450,   0.9459)
(0.9500,   0.9508)
(0.9550,   0.9558)
(0.9600,   0.9607)
(0.9650,   0.9656)
(0.9700,   0.9705)
(0.9750,   0.9754)
(0.9800,   0.9803)
(0.9850,   0.9853)
(0.9900,   0.9902)
(0.9950,   0.9951)
(1.0000,    1.0000)
};
\addlegendentry{{}{Without Speed}}

\addplot+[
  mark = {none},
  solid, thick, pastelBlue
] coordinates {
  (0,    0.0047)
(0.0050,    0.0221)
(0.0100,    0.0328)
(0.0150,    0.0421)
(0.0200,    0.0508)
(0.0250,    0.0589)
(0.0300,    0.0667)
(0.0350,    0.0741)
(0.0400,    0.0814)
(0.0450,    0.0885)
(0.0500,    0.0953)
(0.0550,    0.1021)
(0.0600,    0.1087)
(0.0650,    0.1153)
(0.0700,    0.1217)
(0.0750,    0.1280)
(0.0800,    0.1343)
(0.0850,    0.1404)
(0.0900,    0.1466)
(0.0950,    0.1526)
(0.1000,    0.1586)
(0.1050,    0.1645)
(0.1100,    0.1704)
(0.1150,    0.1762)
(0.1200,    0.1820)
(0.1250,    0.1878)
(0.1300,    0.1935)
(0.1350,    0.1991)
(0.1400,    0.2048)
(0.1450,    0.2104)
(0.1500,    0.2159)
(0.1550,    0.2214)
(0.1600,    0.2269)
(0.1650,    0.2324)
(0.1700,    0.2378)
(0.1750,    0.2433)
(0.1800,    0.2486)
(0.1850,    0.2540)
(0.1900,    0.2593)
(0.1950,    0.2647)
(0.2000,    0.2699)
(0.2050,    0.2752)
(0.2100,    0.2805)
(0.2150,    0.2857)
(0.2200,    0.2909)
(0.2250,    0.2961)
(0.2300,    0.3013)
(0.2350,    0.3064)
(0.2400,    0.3116)
(0.2450,    0.3167)
(0.2500,    0.3218)
(0.2550,    0.3269)
(0.2600,    0.3319)
(0.2650,    0.3370)
(0.2700,    0.3420)
(0.2750,    0.3471)
(0.2800,    0.3521)
(0.2850,    0.3571)
(0.2900,    0.3621)
(0.2950,    0.3670)
(0.3000,    0.3720)
(0.3050,    0.3769)
(0.3100,    0.3819)
(0.3150,    0.3868)
(0.3200,    0.3917)
(0.3250,    0.3966)
(0.3300,    0.4015)
(0.3350,    0.4064)
(0.3400,    0.4112)
(0.3450,    0.4161)
(0.3500,    0.4209)
(0.3550,    0.4258)
(0.3600,    0.4306)
(0.3650,    0.4354)
(0.3700,    0.4402)
(0.3750,    0.4450)
(0.3800,    0.4498)
(0.3850,    0.4546)
(0.3900,    0.4594)
(0.3950,    0.4641)
(0.4000,    0.4689)
(0.4050,    0.4736)
(0.4100,    0.4783)
(0.4150,    0.4831)
(0.4200,    0.4878)
(0.4250,    0.4925)
(0.4300,    0.4972)
(0.4350,    0.5019)
(0.4400,    0.5066)
(0.4450,    0.5113)
(0.4500,    0.5159)
(0.4550,    0.5206)
(0.4600,    0.5253)
(0.4650,    0.5299)
(0.4700,    0.5346)
(0.4750,    0.5392)
(0.4800,    0.5438)
(0.4850,    0.5485)
(0.4900,    0.5531)
(0.4950,    0.5577)
(0.5000,    0.5623)
(0.5050,    0.5669)
(0.5100,    0.5715)
(0.5150,    0.5761)
(0.5200,    0.5806)
(0.5250,    0.5852)
(0.5300,    0.5898)
(0.5350,    0.5943)
(0.5400,    0.5989)
(0.5450,    0.6035)
(0.5500,    0.6080)
(0.5550,    0.6125)
(0.5600,    0.6171)
(0.5650,    0.6216)
(0.5700,    0.6261)
(0.5750,    0.6306)
(0.5800,    0.6352)
(0.5850,    0.6397)
(0.5900,    0.6442)
(0.5950,    0.6487)
(0.6000,    0.6532)
(0.6050,    0.6576)
(0.6100,    0.6621)
(0.6150,    0.6666)
(0.6200,    0.6711)
(0.6250,    0.6755)
(0.6300,    0.6800)
(0.6350,    0.6845)
(0.6400,    0.6889)
(0.6450,    0.6934)
(0.6500,    0.6978)
(0.6550,    0.7023)
(0.6600,    0.7067)
(0.6650,    0.7111)
(0.6700,    0.7156)
(0.6750,    0.7200)
(0.6800,    0.7244)
(0.6850,    0.7288)
(0.6900,    0.7332)
(0.6950,    0.7376)
(0.7000,    0.7420)
(0.7050,    0.7464)
(0.7100,    0.7508)
(0.7150,    0.7552)
(0.7200,    0.7596)
(0.7250,    0.7640)
(0.7300,    0.7684)
(0.7350,    0.7728)
(0.7400,    0.7771)
(0.7450,    0.7815)
(0.7500,    0.7859)
(0.7550,    0.7902)
(0.7600,    0.7946)
(0.7650,    0.7989)
(0.7700,    0.8033)
(0.7750,    0.8076)
(0.7800,    0.8120)
(0.7850,    0.8163)
(0.7900,    0.8206)
(0.7950,    0.8250)
(0.8000,    0.8293)
(0.8050,    0.8336)
(0.8100,    0.8380)
(0.8150,    0.8423)
(0.8200,    0.8466)
(0.8250,    0.8509)
(0.8300,    0.8552)
(0.8350,    0.8595)
(0.8400,    0.8638)
(0.8450,    0.8681)
(0.8500,    0.8724)
(0.8550,    0.8767)
(0.8600,    0.8810)
(0.8650,    0.8853)
(0.8700,    0.8896)
(0.8750,    0.8939)
(0.8800,    0.8981)
(0.8850,    0.9024)
(0.8900,    0.9067)
(0.8950,    0.9110)
(0.9000,    0.9152)
(0.9050,    0.9195)
(0.9100,    0.9238)
(0.9150,    0.9280)
(0.9200,    0.9323)
(0.9250,    0.9365)
(0.9300,    0.9408)
(0.9350,    0.9450)
(0.9400,    0.9493)
(0.9450,    0.9535)
(0.9500,    0.9577)
(0.9550,    0.9620)
(0.9600,    0.9662)
(0.9650,    0.9705)
(0.9700,    0.9747)
(0.9750,    0.9789)
(0.9800,    0.9831)
(0.9850,    0.9874)
(0.9900,    0.9916)
(0.9950,    0.9958)
(1.0000,    1.0000)
};
\addlegendentry{{}{With Speed}}

\end{axis}
\end{tikzpicture}